\documentclass[11pt,a4paper]{article}
\usepackage[hyperref]{emnlp2020}
\usepackage{times}
\usepackage{latexsym}
\usepackage{booktabs} 
\usepackage{multirow}
\usepackage{url}
\usepackage{amsmath}
\usepackage{amsfonts}
\usepackage{graphicx}
\usepackage{bm}
\usepackage{color}
\usepackage{diagbox}
\usepackage{xspace}
\usepackage{courier}
\usepackage{mathrsfs}
\usepackage{appendix}
\usepackage{float}
\usepackage{framed} 

\usepackage{microtype}

\newcommand\CorefBERT{CorefBERT\xspace}
\newcommand\BASESIZE{$_{\small \textsc{base}}$\xspace}
\newcommand\LARGESIZE{$_{\small \textsc{large}}$\xspace}

\aclfinalcopy 
\def\aclpaperid{315} 

\title{Coreferential Reasoning Learning for Language Representation}
\author{Deming Ye$^{1,2}$, Yankai Lin$^{4}$, Jiaju Du$^{1,2}$, Zhenghao Liu$^{1,2}$,  Peng Li$^{4}$, Maosong Sun$^{1,3}$, Zhiyuan Liu$^{1,3}$\\
$^1$Department of Computer Science and Technology, Tsinghua University, Beijing, China\\
Institute for Artificial Intelligence, Tsinghua University, Beijing, China\\
Beijing National Research Center for Information Science and Technology\\
$^2$State Key Lab on Intelligent Technology and Systems, Tsinghua University, Beijing, China\\
$^3$Beijing Academy of Artificial Intelligence \\
$^{4}$Pattern Recognition Center, WeChat AI, Tencent Inc.\\
\texttt{ydm18@mails.tsinghua.edu.cn}
}

\date{}

\begin{document}
\maketitle
\begin{abstract}

Language representation models such as BERT could effectively capture contextual semantic information from plain text, and have been proved to achieve promising results in lots of downstream NLP tasks with appropriate fine-tuning. 
However, most existing language representation models cannot explicitly handle coreference, which is essential to the coherent understanding of the whole discourse.
To address this issue, we present CorefBERT, a novel language representation model that can capture the coreferential relations in context.
The experimental results show that, compared with existing baseline models,  CorefBERT can achieve significant improvements consistently on various downstream NLP tasks that require coreferential reasoning, while maintaining comparable performance to previous models on other common NLP tasks. The source code and experiment details of this paper can be obtained from \url{https://github.com/thunlp/CorefBERT}.

\end{abstract}

\definecolor{olive}{RGB}{84,130,53}
\definecolor{rel}{RGB}{237,125,49}

\section{Introduction}


Recently, language representation models such as BERT~\citep{BERT} have attracted considerable attention. These models usually conduct self-supervised pre-training tasks over large-scale corpus to obtain informative language representation, which could capture the contextual semantic of the input text. 
Benefiting from this, language representation models have made significant strides in many natural language understanding tasks including natural language inference~\citep{nlibert},  sentiment classification~\citep{sentimentbert}, question answering~\citep{qabert}, relation extraction~\citep{peters-etal-2019-knowledge}, fact extraction and verification~\citep{Zhoujie}, and coreference resolution~\citep{bertforcoref}.




However, existing pre-training tasks, such as masked language modeling, usually only require models to collect local semantic and syntactic information to recover the masked tokens. Hence, language representation models may not well model the long-distance connections beyond sentence boundary in a text, such as coreference. Previous work has shown that the performance of these models is not as good as human performance on the tasks requiring coreferential reasoning~\citep{Lambada, QUOREF}, and they can be further improved on long-text tasks with external coreference information~\citep{Coreflambada,CorefSummary, Transformer-XH}. Coreference occurs when two or more expressions in a text refer to the same entity, which is an important element for a coherent understanding of the whole discourse. For example, for comprehending the whole context of ``\emph{Antoine published The Little Prince in 1943. The book follows a young prince who visits various planets in space.}'', we must realize that \emph{The book} refers to \emph{The Little Prince}. Therefore, resolving coreference is an essential step for abundant higher-level NLP tasks requiring full-text understanding.

\begin{figure*}
    \centering
    \includegraphics[width=\textwidth]{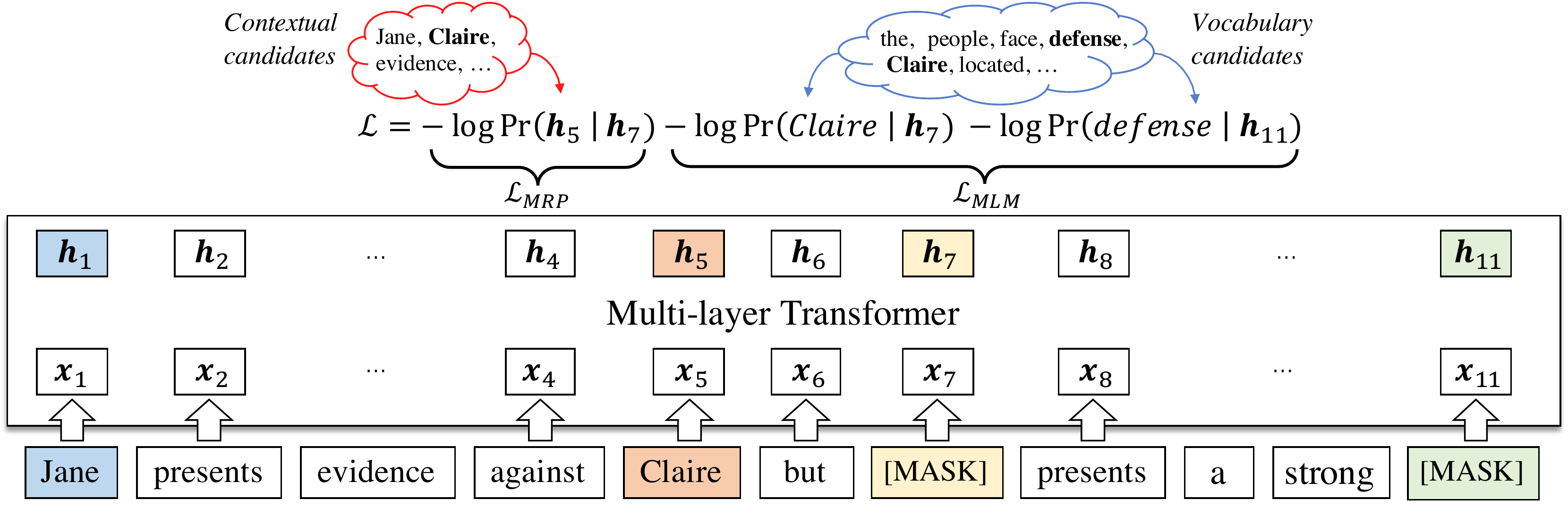}
    \caption{An illustration of CorefBERT's  training process. In this example, the second \emph{Claire} and a common word \emph{defense} are masked. The overall loss of \emph{Claire} consists of the loss of both Mention Reference Prediction (MRP) and Masked Language Modeling (MLM). MRP requires model to select contextual candidates to recover the masked tokens, while MLM asks model to choose from vocabulary candidates. In addition, we also sample some other tokens, such as \emph{defense} in the figure, which is only trained with MLM loss.}
    \label{fig:model}
\end{figure*}

To improve the capability of coreferential reasoning for language representation models, a straightforward solution is to fine-tune these models on supervised coreference resolution data. Nevertheless, on the one hand, we find fine-tuning on existing small coreference datasets cannot improve the model performance on downstream tasks in our preliminary experiments. On the other hand, it is impractical to obtain a large-scale supervised coreference dataset. 

To address this issue, we present \CorefBERT, a language representation model designed to better capture and represent the coreference information. To learn coreferential reasoning ability from large-scale unlabeled corpus, \CorefBERT introduces a novel pre-training task called \emph{Mention Reference Prediction} (MRP). MRP leverages those repeated mentions (e.g., noun or noun phrase) that appear multiple times in the passage to acquire abundant co-referring relations. 
Among the repeated mentions in a passage, MRP applies mention reference masking strategy to mask one or several mentions  and requires model to predict the masked mention's corresponding referents. 
Figure~\ref{fig:model} shows an example of the MRP task, we substitute one of the repeated mentions, \emph{Claire}, with \texttt{[MASK]} and ask the model to find the proper contextual candidate for filling it. 
To explicitly model the coreference information, we further introduce a copy-based training objective to encourage the model to select words from context instead of the whole vocabulary. The internal logic of our method is essentially similar to that of coreference resolution, which aims to find out all the mentions that refer to the masked mentions in a text. Besides, rather than using a context-free word embedding matrix when predicting words from the vocabulary, copying from context encourages the model to generate more context-sensitive representations, which is more feasible to model coreferential reasoning.

We conduct experiments on a suite of downstream tasks which require coreferential reasoning in language understanding, including extractive question answering, relation extraction, fact extraction and verification, and coreference resolution. The results show that \CorefBERT outperforms the vanilla BERT on almost all benchmarks and even strengthens the performance of the strong RoBERTa model. 
To verify the model's robustness, we also evaluate CorefBERT on other common NLP tasks where CorefBERT still achieves comparable results to BERT. It demonstrates that the introduction of the new pre-training task about coreferential reasoning would not impair BERT's ability in common language understanding.

\section{Related Work}


Pre-training language representation models aim to capture language information from the text, which facilitate various downstream NLP applications~\citep{sentenceclassfication,relationextraction,BIDAF}. Early works~\citep{word2vec,glove} focus on learning static word embeddings from the unlabeled corpus, which have the limitation that they cannot handle the polysemy well. Recent years, contextual language representation models pre-trained on large-scale unlabeled corpora have attracted intensive attention and efforts from both academia and industry. SA-LSTM~\citep{Semisuperviedsequencelearning} and ULMFiT~\citep{ULMFiT} pre-trains language models on unlabeled text and perform task-specific fine-tuning. ELMo~\citep{elmo} further employs a bidirectional LSTM-based language model to extract context-aware word embeddings. Moreover, OpenAI GPT~\citep{GPT} and BERT~\citep{BERT} learn pre-trained language representation with Transformer architecture~\citep{Transformer}, achieving state-of-the-art results on various NLP tasks. Beyond them, various improvements on pre-training language representation have been proposed more recently, including (1) designing new pre-trainning tasks or objectives such as SpanBERT~\citep{Spanbert} with span-based learning, XLNet~\citep{XLNET} considering masked positions dependency with auto-regressive loss, MASS~\citep{MASS} and BART~\citep{wang-etal-2019-denoising} with sequence-to-sequence pre-training, ELECTRA~\citep{clark2019electra} learning from replaced token detection with generative adversarial networks and InfoWord~\citep{kong2019mutual} with contrastive learning; 
(2) integrating external knowledge such as factual knowledge in knowledge graphs~\citep{TsinghuaERNIE,peters-etal-2019-knowledge,weijie2019kbert}; and (3) exploring multilingual learning~\citep{conneau2019cross,tan-bansal-2019-lxmert,kondratyuk-straka-2019-75} or multimodal learning~\citep{lu2019vilbert,Sun_2019_ICCV,su2019vl}.  Though existing language representation models have achieved a great success, their coreferential reasoning capability are still far less than that of human beings~\citep{Lambada,QUOREF}. In this paper, we design a mention reference prediction task to enhance language representation models in terms of coreferential reasoning.

Our work, which acquires coreference resolution ability from an unlabeled corpus, can also be viewed as a special form of unsupervised coreference resolution.
Formerly, researchers have made efforts to explore feature-based unsupervised coreference resolution methods~\citep{UnsupervisedModelsforCoreferenceResolution, UnsupervisedRankingModelforEntityCoreferenceResolution}. 
After that, Word-LM~\citep{ASimpleMethodforCommonsenseReasoning} uncovers that it is natural to resolve pronouns in the sentence according to the probability of language models. Moreover, WikiCREM~\citep{wikicrem} 
builds sentence-level unsupervised coreference resolution dataset for learning coreference discriminator. 
However, these methods cannot be directly transferred to language representation models since their task-specific design could  weaken the model's performance on other NLP tasks. To address this issue, we introduce a mention reference prediction objective, complementary to masked language modeling, which could make the obtained coreferential reasoning ability compatible with more downstream tasks.
\section{Methodology}
In this section, we present \CorefBERT, a language representation model, which aims to better capture the coreference information of the text. As illustrated in Figure~\ref{fig:model}, \CorefBERT adopts the deep bidirectional Transformer architecture~\citep{Transformer} and utilizes two training tasks: 

(1) \textbf{Mention Reference Prediction} (MRP) is a novel  training task which is proposed to  enhance coreferential reasoning ability.
MRP utilizes the mention reference masking strategy to mask one of the  repeated mentions and then employs a copy-based training objective to predict the masked tokens by copying from other tokens in the sequence.

(2) \textbf{Masked Language Modeling} (MLM)\footnote{Details of MLM are in the appendix due to space limit.} is proposed  from vanilla BERT~\citep{BERT}, aiming to learn the general language understanding. MLM is regarded as a kind of cloze tasks and aims to predict the missing tokens according to its final contextual representation. Except for MLM, Next Sentence Prediction (NSP) is also commonly used in BERT, but we train our model without the NSP objective since some previous works~\citep{Roberta, Spanbert} have revealed that NSP is not as helpful as expected.

Formally, given a sequence of tokens\footnote{In this paper, tokens are at the subword level.} ${X} = (x_1, x_2, \ldots, x_n)$, we first represent each token by aggregating the corresponding token and position embeddings, and then feeds the input representations into deep bidirectional Transformer to obtain the contextual representations, which is used to compute the loss for pre-training tasks. The overall loss of \CorefBERT is composed of two training losses: the mention reference prediction loss $L_{\text{MRP}}$ and the masked language modeling loss $L_{\text{MLM}}$, which can be formulated as:
\begin{equation}
    \mathcal{L} = \mathcal{L}_{\text{MRP}} +  \mathcal{L}_{\text{MLM}}.
\end{equation}


%

\subsection{Mention Reference Masking}

To better capture the coreference information in the text, we propose a novel masking strategy: mention reference masking, which masks tokens of the repeated mentions in the sequence instead of masking random tokens. We follow a distant supervision assumption: the repeated mentions in a sequence would refer to each other. Therefore, if we mask one of them, the masked tokens would be inferred through its context and  unmasked references. Based on the above strategy and assumption, the CorefBERT model is expected to capture the coreference information in the text for filling the masked token. 

In practice, we regard nouns in the text as mentions. We first use a part-of-speech tagging tool to extract all nouns in the given sequence. 
Then, we cluster the nouns into several groups where each group contains all mentions of the same noun. After that, we select the masked nouns from different groups uniformly. For example, when \emph{Jane} occurs three times and \emph{Claire} occurs two time in the text, all the mentions of \emph{Jane} or \emph{Claire} will be grouped.  Then, we choose one of the groups, and then sample one mention of the selected group.

To maintain the universal language representation ability in \CorefBERT, we utilize both the MLM (masking random word) and MRP (masking mention reference)  in the training process. Empirically, the masked words for MLM and MRP are sampled on a ratio of 4:1. Similar to BERT, $15\%$ of the tokens are sampled for both masking strategies mentioned above, where $80\%$ of them are replaced with a special token \texttt{[MASK]}, 
$10\%$ of them are replaced with random tokens, and $10\%$ of them are unchanged.
We also adopt whole word masking (WWM)~\citep{Spanbert}, which masks all the subwords belong to the masked words or mentions.

\subsection{Copy-based Training Objective}

In order to capture the coreference information of the text, CorefBERT models the correlation among words in the sequence. 
Inspired by copy mechanism~\citep{copynet,copynetMT} in sequence-to-sequence tasks, we introduce a copy-based training objective to require the model to predict missing tokens of the masked mention by copying the unmasked tokens in the context. Since the masked tokens would be copied from context, low-frequency tokens, such as proper nouns, could be well processed to some extent.
Moreover, through copying mechanism, the \CorefBERT model could explicitly capture the relations between the masked mention and its referring mentions, therefore, to obtain the coreference information in the context.

Formally, we first encode the given input sequence ${X} = (x_1,\ldots, x_{n})$ into hidden states $\bm{H} = (\bm{h}_1,\ldots, \bm{h}_{n})$ via multi-layer Transformer~\citep{Transformer}. The probability of recovering the masked token $x_i$ by copying from $x_j$ is defined as:
\begin{equation}
\Pr(x_j  | x_i) = \frac{\exp( (\bm{V}\odot\bm{h}_j )^T\bm{h}_i)}{\sum_{x_k \in X}\exp((\bm{V}\odot\bm{h}_k)^T\bm{h}_i)},
\end{equation}
where $\odot$ denotes element-wise product function and $\bm{V}$ is a trainable parameter to measure the importance of each dimension for token's similarity.

Moreover, since we split a word into several word pieces as BERT does and we adopt whole word masking strategy for MRP, we need to extend our copy-based objective into word-level. To this end, we apply the token-level copy-based training objective on both start and end tokens of the masked word, because the representations of these two tokens could typically cover the major information of the whole word~\citep{e2ecoref2,neuralsemanticrolelabeling}. For a masked noun $w_i$ consisting of a sequence of tokens $(x_s^{(i)},\ldots, x_{t}^{(i)})$, we recover $w_i$ by copying its referring context word, and define the probability of choosing word $w_j$ as:
\begin{equation}
\Pr(w_j  | w_i) = \Pr(x_s^{(j)}  | x_s^{(i)})  \times \Pr(x_t^{(j)}  | x_t^{(i)}).
\end{equation}

A masked noun possibly has multiple referring words in the sequence, for which we collectively maximize the similarity of all referring words. It is an approach widely used in question answering ~\citep{ASReader, NeuralCascades, sharednorm} designed to handle multiple answers. Finally, we define the loss of Mention Reference Prediction (MRP) as:
\begin{equation}
\mathcal{L}_{\text{MRP}} = - \sum_{w_i \in {M} } \log \sum_{w_j \in C_{w_i}} \Pr(w_j|w_i),
\end{equation}
where $M$ is the set of all masked mentions for mention reference masking, and  $C_{w_i}$ is the set of all corresponding words of word  $w_i$.




\section{Experiment}
In this section, we first introduce the training details of  \CorefBERT. After that, we present the fine-tuning results on a comprehensive suite of tasks, including extractive question answering, document-level relation extraction, fact extraction and verification, coreference resolution, and eight tasks in the GLUE benchmark.

\subsection{Training Details}
Since training \CorefBERT from scratch would be time-consuming, we initialize the parameters of \CorefBERT with BERT  released by Google\footnote{{https://github.com/google-research/bert}}, which is also used as our baselines on  downstream tasks. 
Similar to previous language representation models~\citep{BERT, Spanbert}, we adopt English Wikipeida\footnote{{https://en.wikipedia.org}} as our training corpus, which contains about 3,000M tokens.  We employ {spaCy}\footnote{{https://spacy.io}} for part-of-speech-tagging on the corpus. 
We train CorefBERT with contiguous sequences of up to $512$ tokens, and randomly shorten the input sequences with $10\%$ probability in training.   To verify the effectiveness of our method for the language representation model trained with tremendous corpus, we also train CorefBERT initialized with RoBERTa\footnote{{https://github.com/pytorch/fairseq}}, referred as CorefRoBERTa.
Additionally, we follow the pre-training hyper-parameters used in BERT, and adopt Adam optimizer~\citep{Adam}  with batch size of $256$.  Learning rate of $5\times 10^{-5}$ is used for the base model and $1\times10^{-5}$ is used for the large model. The optimization runs $33$k steps, 
where the learning rate is warmed-up over the first 20\% steps and then linearly decayed.
The pre-training process consumes $1.5$ days for base model and $11$ days for large model with 8 RTX 2080 Ti GPUs in mixed precision. We search the ratio of MRP loss and MLM loss in $1$:$1$, $1$:$2$ and $2$:$1$, and find the ratio of $1$:$1$ achieves the best result. Beyond this, training details for downstream tasks are shown in the appendix.

\subsection{Extractive Question Answering}

Given a question and passage, the extractive question answering task aims to select spans in passage to answer the question. We first evaluate models on  Questions Requiring Coreferential Reasoning dataset (QUOREF)~\citep{QUOREF}.
Compared to previous reading comprehension benchmarks, QUOREF is more challenging as $78$\% of the questions in QUOREF cannot be answered without coreference resolution. In  this case, it can be an effective tool to examine the coreferential reasoning capability of question answering models. 

We also adopt the MRQA, a dataset not specially designed for examining  coreferential reasoning capability, which involves paragraphs from different sources and questions with manifold styles. Through MRQA, we hope to evaluate the performance of our model in various domains.
We use six benchmarks of MRQA, including SQuAD~\citep{rajpurkar2016squad}, NewsQA~\citep{newsqa}, SearchQA~\citep{searchqa}, TriviaQA~\citep{triviaqa}, HotpotQA~\citep{hotpotqa}, and Natural Questions (NaturalQA)~\citep{naturalqa}. 
Since MRQA does not provide a public test set, we randomly split the development set into two halves to generate new validation and test sets.


\vspace{-0.1em}
\paragraph{Baselines} For QUOREF, we compare our CorefBERT  with four baseline models: (1) \textbf{QANet}~\citep{QANET} combines self-attention mechanism with the convolutional neural network, which achieves the best performance to date without pre-training; (2) \textbf{QANet+BERT} adopts BERT representation as an additional input feature into QANet; (3) \textbf{BERT}~\citep{BERT}, simply fine-tunes BERT for extractive question answering.   
We further design two components accounting for coreferential reasoning and multiple answers, by which we obtain stronger BERT baselines;
(4) \textbf{RoBERTa-MT}  trains RoBERTa on CoLA, SST2, SQuAD datasets before on QUOREF. For MRQA, we compare CorefBERT to vanilla BERT with the same question answering framework.

\begin{table}[!t]
\small
\centering
\begin{tabular}{l c c c c}
\toprule

\multirow{2}{*}{Model}   &\multicolumn{2}{c}{Dev}   &  \multicolumn{2}{c}{Test}  \\
                        &  EM        & F1            & EM        & F1 \\
\midrule
QANet$^*$ & 34.41& 38.26& 34.17& 38.90 \\
QANet+BERT\BASESIZE $^*$ & 43.09 & 47.38& 42.41& 47.20 \\
BERT\BASESIZE $^*$ & 58.44 & 64.95& 59.28& 66.39 \\
BERT\BASESIZE  & 61.29 &  67.25   & 61.37  & 68.56  \\
CorefBERT$_{\text{Base}}$  & \bf{66.87} & \bf{72.27} & \bf{66.22}& \bf{72.96} \\
\midrule
BERT\LARGESIZE  & 67.91 & 73.82 & 67.24  & 74.00  \\
CorefBERT\LARGESIZE  & \bf{70.89} & \bf{76.56} & \bf{70.67} & \bf{76.89}  \\
\midrule
RoBERTa-MT$^+$  & 74.11 & 81.51 & 72.61 & 80.68 \\
RoBERTa\LARGESIZE &  74.15 & 81.05 &  75.56 & 82.11 \\
CorefRoBERTa\LARGESIZE  & \bf{74.94} & \bf{81.71} & \bf{75.80} & \bf{82.81} \\
\bottomrule
\end{tabular}
\caption{Results on QUOREF measured by exact match (EM) and F1. Results with $^*$, $^+$ are from \citet{QUOREF} and official leaderboard respectively. }
\label{tab:QUOREF_results}
\end{table}

\vspace{-0.1em}
\paragraph{Implementation Details} Following BERT's setting~\citep{BERT}, given the question $Q = (q_1, q_2, \ldots, q_m)$ and the passage $P = (p_1, p_2, \ldots, p_n)$, we represent them as a sequence $\bm{X} = (\text{[CLS]},q_1, q_2, \ldots, q_m,\text{[SEP]},$ $p_1, p_2,\ldots, p_n,\text{[SEP]})$, feed the sequence $\bm{X}$ into the pre-trained encoder and train two classifiers on the top of it to seek answer's start and end positions simultaneously. For MRQA, CorefBERT maintains the same framework as BERT. For QUOREF, we further employ two extra components to process multiple mentions of the answers: (1) Spurred by the idea from MTMSN~\citep{MTMSN} in handling the problem of multiple answer spans, we utilize the representation of \textup{[CLS]} to predict the number of answers. After that, we first selects the answer span of the current highest scores, then continues to choose that of the second-highest score with no overlap to previous spans, until reaching the predicted answer number. 
(2) When answering a question from QUOREF, the relevant mention could possibly be a pronoun, so we attach a reasoning Transformer layer for pronoun resolution before the span boundary classifier.

\begin{table*}[!t]
\small
\centering
\setlength{\tabcolsep}{8pt}
\begin{tabular}{lccccccc}
\toprule

 Model &    {SQuAD} & {NewsQA}  & {TriviaQA} & {SearchQA} &{HotpotQA} &{NaturalQA} & Average  \\
\midrule
BERT\BASESIZE  & 88.4& 66.9 &68.8  & 78.5  & 74.2  & 75.6 & 75.4 \\
CorefBERT\BASESIZE  &   \bf{89.0}& \bf{69.5}& \bf{70.7} & \bf{79.6} & \bf{76.3} & \bf{77.7} & \bf{77.1}\\ 
\midrule
BERT\LARGESIZE  & 91.0& 69.7 & 73.1  & 81.2  & 77.7  & 79.1 & 78.6 \\
CorefBERT\LARGESIZE  &\bf{91.8} & \bf{71.5} &  \bf{73.9} & \bf{82.0}  & \bf{79.1}  & \bf{79.6} &  \bf{79.6}\\
\bottomrule
\end{tabular}
\caption{Performance (F1)  on six MRQA extractive question answering benchmarks.} 
\label{tab:MRQA_result}
\end{table*}




\vspace{-0.1em}
\paragraph{Results} Table~\ref{tab:QUOREF_results} shows the performance on QUOERF. Our
adapted BERT\BASESIZE surpasses original BERT by about $2\%$ in EM and F1 score, indicating the effectiveness of the added reasoning layer and multi-span prediction module. CorefBERT\BASESIZE and CorefBERT\LARGESIZE  exceeds our adapted BERT\BASESIZE and BERT\LARGESIZE by  $4.4\%$ and $2.9\%$ F1 respectively. 
Leaderboard results are shown in the appendix. Based on the TASE framework~\citep{TASE}, the model with CorefRoBERTa achieves a new state-of-the-art with about $1\%$ EM improvement compared to the model with RoBERTa.
We also show four case studies in the appendix, which indicate that through reasoning over mentions, \CorefBERT could aggregate information to answer the question requiring coreferential reasoning.


Table~\ref{tab:MRQA_result} further shows that the effectiveness of \CorefBERT is consistent in six datasets of the MRQA shared task besides QUOREF. Though the MRQA shared task is not designed for coreferential reasoning, CorefBERT still achieves averagely over $1\%$ F1 improvement on all of the six datasets, especially on NewsQA and HotpotQA. In NewsQA, $20.7\%$ of the answers can only be inferred by synthesizing information distributed across multiple sentences.  In HotpotQA,  $63\%$ of the answers need to be inferred through either bridge entities or checking multiple properties in different positions. It demonstrates that coreferential reasoning is an essential ability in question answering.


\subsection{Relation Extraction}

Relation extraction (RE) aims to extract the relationship between two entities in a given text. We evaluate our model on DocRED~\citep{DocRED}, a challenging document-level RE dataset which requires the model to extract relations between entities by synthesizing information from all the mentions of them after reading the whole document. 
DocRED requires a variety of reasoning types
, where $17.6$\% of the relational facts need to be uncovered through coreferential reasoning. 

\vspace{-0.1em}
\paragraph{Baselines} We compare our model with the following baselines for document-level relation extraction: (1) \textbf{CNN} / \textbf{LSTM} / \textbf{BiLSTM} / \textbf{BERT}. CNN~\citep{CNNRE}, LSTM~\citep{LSTM}, bidirectional LSTM (BiLSTM)~\citep{BiLSTM}, BERT~\citep{BERT} are widely adopted as text encoders in relation extraction tasks. 
With these encoders, \citet{DocRED} generates representations of entities for further predicting of the relationships between entities.
(2) \textbf{ContextAware}~\citep{ContextAware} takes relations' interaction into account, which demonstrates that other relations in the context are beneficial for target relation prediction. 
(3) \textbf{BERT-TS}~\citep{DocREDBert} applies a two-step prediction to deal with the large number of irrelevant entities, which first predicts whether two entities have a relationship and then predicts the specific relation.
(4) \textbf{HinBERT}~\citep{HinBERT} proposes a hierarchical inference network to  aggregate the inference information with different granularity.

\begin{table}[!t]
\small
\centering
\begin{tabular}{l c c c c}
\toprule

\multirow{2}{*}{Model}   &\multicolumn{2}{c}{Dev}   &  \multicolumn{2}{c}{Test}  \\
                        &  IgnF1        & F1            & IgnF1        & F1 \\
\midrule
CNN$^*$ & 41.58 &  43.45 &  40.33 &  42.26 \\
LSTM$^*$ & 48.44 &50.68 & 47.71 & 50.07\\
BiLSTM$^*$ & {48.87}  & 50.94 & 50.26 & 51.06 \\
ContextAware$^*$ & 48.94 & 51.09 & 48.40  & 50.70 \\  
\midrule
BERT-TS\BASESIZE $^+$  & - & 54.42 & -& 53.92\\
HINBERT\BASESIZE $^{\#}$ & 54.29 &56.31 & 53.70 & 55.60 \\
BERT\BASESIZE  &54.63 & 56.77 & 53.93 & 56.27 \\
CorefBERT\BASESIZE & \bf{55.32} & \bf{57.51} & \bf{54.54} & \bf{56.96} \\
\midrule
BERT\LARGESIZE  & 56.51 & 58.70 & 56.01 & 58.31 \\
CorefBERT\LARGESIZE & \bf{56.82} & \bf{59.01} & \bf{56.40} & \bf{58.83} \\
\midrule
RoBERTa\LARGESIZE & 57.19 & {59.40} & 57.74 & 60.06\\
CorefRoBERTa\LARGESIZE & \bf{57.35} & \bf{59.43} & \bf{57.90} & \bf{60.25} \\
\bottomrule
\end{tabular}
\caption{Results on DocRED measured by micro ignore F1 (IgnF1) and micro F1. IgnF1 metrics ignores the relational facts shared by the training and dev/test sets. Results with $^*$, $^+$, $^{\#}$ are from \citet{DocRED}, \citet{DocREDBert}, and \citet{HinBERT} respectively.}
\label{tab:DocRED_result}
\end{table}

\vspace{-0.1em}
\paragraph{Results} Table~\ref{tab:DocRED_result} shows the performance on DocRED. The BERT\BASESIZE model we implemented with mean-pooling entity representation and hyperparameter tuning\footnote{Details are in the appendix due to space limit.}  performed better than previous RE models with BERT\BASESIZE size, which provides a stronger baseline. CorefBERT\BASESIZE outperforms BERT\BASESIZE model by $0.7$\% F1. CorefBERT\LARGESIZE beats BERT\LARGESIZE by $0.5\%$ F1. 
We also show a case study in the appendix, which further proves that considering coreference information of text is helpful for exacting relational facts from documents.

\subsection{Fact Extraction and Verification}
Fact extraction and verification aim to verify deliberately fabricated claims with trust-worthy corpora. We evaluate our model  on a large-scale public fact verification dataset FEVER~\citep{Fever}. FEVER consists of $185,455$ annotated claims with all Wikipedia documents.

\vspace{-0.1em}
\paragraph{Baselines}
We compare our model with four BERT-based fact verification models: (1) \textbf{BERT Concat}~\citep{Zhoujie} concatenates all of the evidence pieces and the claim to predict the claim label; (2) \textbf{SR-MRS}~\citep{SR-MRS} employs hierarchical BERT retrieval to improve the performance; (3) \textbf{GEAR}~\citep{Zhoujie} constructs an evidence graph and conducts a graph attention network for jointly reasoning over several evidence pieces; (4) \textbf{KGAT}~\citep{Zhenghao}  conducts a fine-grained graph attention network with kernels.

\vspace{-0.1em}
\paragraph{Results}
Table~\ref{tab:Fever_result} shows the performance on FEVER. KGAT with CorefBERT\BASESIZE outperforms KGAT with BERT\BASESIZE by $0.4$\% FEVER score. KGAT with CorefRoBERTa\LARGESIZE gains $1.9\%$ FEVER score improvement compared to the model with RoBERTa\LARGESIZE, and arrives at a new state-of-the-art on FEVER benchmark. It again demonstrates the effectiveness of our model. CorefBERT, which incorporates coreference information in distant-supervised pre-training, contributes to verify if the claim and evidence discuss about the same mentions, such as a person or an object.



\subsection{Coreference Resolution}

Coreference resolution aims to link referring expressions that evoke the same discourse entity. We examine models'  coreference resolution ability under the setting that all mentions have been detected. 
We evaluate models on several widely-used datasets, including GAP~\citep{GAP},  DPR~\citep{DPR}, WSC~\citep{WSC},  Winogender~\citep{Winogender} and PDP~\citep{PDP}.  
These datasets provide two sentences where the former has two or more mentions and the latter contains an ambiguous pronoun. 
It is required that the ambiguous pronoun should be connected to the right mention.

\vspace{-0.1em}
\paragraph{Baselines}
We  compare our model with two coreference resolution models: 
(1) \textbf{BERT-LM}~\citep{ASimpleMethodforCommonsenseReasoning} substitutes the pronoun with \texttt{[MASK]} and uses language model to compute the probability of recovering the mention candidates; (2) \textbf{WikiCREM}~\citep{wikicrem} generates GAP-like sentences automatically and trains BERT by minimizing the perplexity of correct mentions on these sentences. Finally, the model is fine-tuned on supervised datasets. Benefiting from the augmented data, WikiCREM achieves state-of-the-art in sentence-level coreference resolution. For BERT-LM and \CorefBERT, we adopt the same data split and the same training method on supervised datasets as those of WikiCREM to the benefit of a fair comparison.





\begin{table}[!t]
 \small
 \centering
 \begin{tabular}{l c c }
 \toprule

 \multirow{1}{*}{Model}   &  LA        & FEVER\\
 \midrule
 BERT Concat$^*$ & 71.01& 65.64\\
 GEAR$^*$    &71.60 &67.10\\
 SR-MRS$^+$     &72.56& 67.26\\
 KGAT (BERT\BASESIZE) $^{\#}$& 72.81& 69.40\\
 KGAT (CorefBERT\BASESIZE) &\bf{72.88} & \bf{69.82}\\
 \midrule
 KGAT (BERT\LARGESIZE) $^{\#}$& 73.61 & 70.24\\
 KGAT (CorefBERT\LARGESIZE) &\bf{74.37} & \bf{70.86}\\
 \midrule
 KGAT (RoBERTa\LARGESIZE) $^{\#}$& 74.07& 70.38\\
 KGAT (CorefRoBERTa$_{\text{Large}}$) &\bf{75.96} & \bf{72.30}\\
\bottomrule
 \end{tabular}
 \caption{Results on FEVER test set measured by label accuracy (LA) and FEVER. The FEVER score evaluates the model performance and considers whether the golden evidence is provided. Results with $^*$, $^+$, $^{\#}$ are from \citet{Zhoujie}, \citet{SR-MRS} and \citet{Zhenghao} respectively. }
 \label{tab:Fever_result}
 \end{table}

\begin{table}[!t]
\small
\centering
\begin{tabular}{l c c c c c c}
\toprule
{Model}  & GAP & DPR  & WSC  & WG & PDP \\
\midrule
BERT-LM\BASESIZE & 75.3 & 75.4 & 61.2   &  68.3 & 76.7 \\
CorefBERT\BASESIZE & \bf{75.7} &  \bf{76.4} & \bf{64.1}&  \bf{70.8}  & \bf{80.0}\\
\midrule
BERT-LM\LARGESIZE$^*$  & 76.0 & 80.1 & 70.0   &  78.8 & 81.7 \\
WikiCREM\LARGESIZE$^*$ & \bf{78.0} & {84.8} & 70.0 & 76.7  & {86.7}\\
CorefBERT\LARGESIZE & 76.8 &  \bf{85.1}& \bf{71.4}&  \bf{80.8}  & \bf{90.0}\\
\midrule
RoBERTa-LM\LARGESIZE  & \bf{77.8} & 90.6 & \bf{83.2}   &  77.1 & 93.3 \\
CorefRoBERTa\LARGESIZE & \bf{77.8} &  \bf{92.2}& \bf{83.2}&  \bf{77.9}  & \bf{95.0}\\
\bottomrule
\end{tabular}
\caption{Results on coreference resolution test sets. Performance on GAP is measured by F1, while scores on the others are given in accuracy.  WG: Winogender. Results with $^*$ are from \citet{wikicrem}. }
\label{tab:Coref_result}
\end{table}

\begin{table*}[!t]
\small
\centering
\begin{tabular}{l c c c c c c c c c}
\toprule

Model &   MNLI-(m/mm)  & QQP & QNLI & SST-2 & CoLA & STS-B & MRPC & RTE & Average \\
\midrule
BERT\BASESIZE &  84.6/83.4 &71.2  & 90.5 & 93.5  & 52.1 & 85.8 &88.9 & 66.4 & 79.6\\
CorefBERT\BASESIZE &  84.2/83.5 &71.3  & 90.5 & 93.7  &51.5 & 85.8 &89.1 & 67.2 & 79.6\\
\midrule
BERT\LARGESIZE &   86.7/85.9  &72.1  & 92.7 & 94.9  & 60.5 & 86.5 &89.3 & 70.1 & 81.9\\
CorefBERT\LARGESIZE &  86.9/85.7 &71.7  & 92.9 & 94.7  & 62.0 & 86.3 & 89.3 & 70.0 & 82.2\\
\bottomrule
\end{tabular}
\caption{Test set performance metrics on GLUE benchmarks. 
Matched/mistached accuracies are reported for MNLI; F1 scores are reported for QQP and MRPC, Spearmanr correlation is reported for STS-B; Accuracy scores are reported for the other tasks.} 
\label{tab:GLUE_result}  
\end{table*}

\begin{table*}[!t]
\small
\centering
\begin{tabular}{lcccccccc}
\toprule

Model &  {QUOREF} & {SQuAD}& {NewsQA}  & {TriviaQA} & {SearchQA} &{HotpotQA} &{NaturalQA}  & {DocRED} \\
\midrule
BERT\BASESIZE & 67.3 & 88.4& 66.9 &68.8  & 78.5  & 74.2  & 75.6 & 56.8 \\
\,-NSP  &70.6  & 88.7 & 67.5 & 68.9 & 79.4  & 75.2 & 75.4 & 56.7 \\
\,-NSP, +WWM  & 70.1 &88.3 & 69.2 & 70.5&  \bf{79.7}& 75.5 & 75.2 & 57.1 \\
\,-NSP, +MRM  & 70.0 & 88.5 & 69.2&  70.2& 78.6 &75.8 & 74.8 & 57.1\\
CorefBERT\BASESIZE  & \bf{72.3} &  \bf{89.0} & \bf{69.5} & \bf{70.7} & 79.6 & \bf{76.3} & \bf{77.7} & \bf{57.5}\\ 


\bottomrule
\end{tabular}
\caption{Ablation study. Results are F1 scores on development set for QUOREF and DocRED, and on test set for others. CorefBERT\BASESIZE combines ``-NSP, +MRM'' scheme and copy-based training objective. } 
\label{tab:Abalation_result}
\end{table*}

\vspace{-0.1em}
\paragraph{Results}
Table~\ref{tab:Coref_result} shows the performance on the test set of the above coreference resolution dataset. Our CorefBERT model significantly outperforms BERT-LM, which demonstrates that the intrinsic coreference resolution ability of  CorefBERT has been enhanced by involving the mention reference prediction training task. Moreover, it achieves comparable performance with state-of-the-art baseline WikiCREM. Note that, WikiCREM is specially designed for sentence-level coreference resolution and is not suitable for other NLP tasks. On the contrary, the coreferential reasoning capability of CorefBERT can be transferred to other NLP tasks. 

\subsection{GLUE}

The Generalized Language Understanding Evaluation  dataset (GLUE)~\citep{GLUE} is designed to evaluate and analyze the performance of models across a diverse range of existing natural language understanding tasks. We evaluate \CorefBERT on the main GLUE benchmark used in BERT.

\vspace{-0.1em}
\paragraph{Implementation Details}
Following BERT's setting, we add \textup{[CLS]} token in front of the input sentences, and extract its representation on the top layer as the whole sentence or sentence pair's representation for classification or regression. 

\vspace{-0.1em}
\paragraph{Results}
Table \ref{tab:GLUE_result} shows the performance on GLUE. We notice that CorefBERT achieves comparable results to BERT. Though GLUE does not require much coreference resolution ability due to its attributes, the results prove that our masking strategy and auxiliary training objective would not weaken the performance on generalized language understanding tasks.

\section{Ablation Study}

In this section, we explore the effects of the Whole Word Masking (WWM), Mention Reference Masking (MRM), Next Sentence Prediction (NSP) and copy-based training objective using several benchmark datasets. We continue to train Google's released BERT\BASESIZE on the same Wikipedia corpus with different strategies. As shown in Table~\ref{tab:Abalation_result}, we have the following observations: (1) Deleting NSP training task triggers a better performance on almost all tasks. The conclusion is consistent with that of RoBERTa~\citep{Roberta}; (2) MRM scheme usually achieves parity with WWM scheme except on SearchQA, and both of them outperform the original subword masking scheme especially on NewsQA (averagely +$1.7\%$ F1) and TriviaQA (averagely +$1.5\%$ F1); (3) On the basis of MRM scheme, our copy-based training objective explicitly requires  model to look for mention's referents in the context, which could adequately consider the coreference information of the sequence. \CorefBERT takes advantage of the objective and further improves the performance, with a substantial gain (+$2.3\%$ F1) on QUOREF.



\section{Conclusion and Future Work}

In this paper, we present a language representation model named \CorefBERT, which is trained on a novel task, Mention Reference Prediction (MRP),  for strengthening the coreferential reasoning ability of BERT. 
Experimental results on several downstream NLP tasks show that our \CorefBERT significantly outperforms BERT by considering the coreference information within the text and even improve the performance of the strong RoBERTa model. In the future, there are several prospective research directions: (1) We introduce a distant supervision (DS) assumption in our MRP training task. However, the automatic labeling mechanism inevitably accompanies with the wrong labeling problem and it is still an open problem to mitigate the noise.
(2) The DS assumption does not consider  pronouns in the text, while  pronouns play an important role in coreferential reasoning. Hence, it is worth developing a novel strategy such as self-supervised learning to further consider the pronoun.


\section{Acknowledgement}
This work is supported by the National Key R\&D Program of China (2020AAA0105200), Beijing Academy of Artificial Intelligence (BAAI) and the NExT++ project from the National Research Foundation, Prime Minister’s Office, Singapore under its IRC@Singapore Funding Initiative.

\bibliography{emnlp2020}
\bibliographystyle{acl_natbib}

\appendix

\section*{Appendices}
\section{Masked Language Modeling (MLM)} 
MLM is regarded as a kind of cloze tasks and aims to predict the missing tokens according to its contextual representation. In our work, $15$\% of the tokens in  input sequence are sampled as the missing tokens. Among them, $80$\% are replaced with a special token \texttt{[MASK]}, $10$\% are replaced with random tokens and $10$\% are unchanged. The task aims to predict original tokens from corrupted input.

\section{Leaderboard Results on QUOREF } 

TASE~\citep{TASE} converts the multi-span prediction problem as a sequence tagging problem, which substantially improves the model's ability in terms of handling multi-span answer. Though the study of TASE and our \CorefBERT are conducted in the same period, we still run TASE with CorefRoBERTa encoder. As Table~\ref{quorefleaderboard} shows, the performance of TASE with CorefRoBERTa encoder gains about $1\%$ EM improvement compared to that with RoBERTa encoder, which demonstrates the effectiveness of \CorefBERT for different question answering frameworks.

\begin{table}[h]
\small
\centering
\begin{tabular}{l c c c c}
\toprule
Model                              & EM        & F1 \\
\midrule
XLNet~\citep{QUOREF} &  61.88 & 71.51 \\ 
RoBERTa-MT  & 72.61 & 80.68 \\
CorefRoBERTa\LARGESIZE  &  {75.80} & {82.81} \\
TASE (RoBERTa)~\citep{TASE} &  79.66 & 86.13 \\
TASE (CorefRoBERTa) & \bf{80.61} & \bf{86.70} \\
\bottomrule
\end{tabular}
\caption{Leaderboard results on QUOREF test set.}
\label{quorefleaderboard}
\end{table}

\begin{table}[!t]
    \begin{center}
    \begin{tabular}{p {0.48\textwidth}}
    \toprule
(1) Q: Whose uncle trains the asthmatic boy? \\    
Paragraph: {\color{olive} [1]} {\color{rel} \bf \texttt{Barry Gabrewski}} is an asthmatic boy ... {\color{olive} [2]} {\color{rel} \bf \texttt{Barry}} wants to learn the martial arts, but is rejected by the arrogant dojo owner Kelly Stone for being too weak. {\color{olive} [3]}
Instead, {\color{rel} \bf \texttt{he}} is taken on as a student by an old Chinese man called {\color{blue}\bf \textit{Mr. Lee}}, {\color{red} \bf Noreen}'s sly uncle. {\color{olive} [4]}  {\color{blue}\bf \textit{Mr. Lee}}  finds creative ways to teach Barry to defend himself from his bullies.  \\
\midrule
(2) Q: Which composer produced String Quartet No. 2? \\
Paragraph: {\color{olive} [1]} {\color{red} \bf Tippett}'s Fantasia on a Theme of Handel for piano and orchestra was performed at the Wigmore Hall in March 1942, with {\color{blue}\bf \textit{Sellick}} again the soloist, and the same venue saw the premiere of {\color{rel} \bf \texttt{the composer}}'s String Quartet No. 2 a year later.  ...  {\color{olive} [2]} In 1942, Schott Music began to publish {\color{red} \bf Tippett}'s works, establishing an association that continued until the end of the {\color{rel} \bf \texttt{the composer}}'s life. \\

\midrule

(3) Q: What is the first name of the person who lost her beloved husband only six months earlier? \\
Pargraph: {\color{olive} [1]} Robert and {\color{blue}\bf \textit{Cathy}} Wilson are a timid married couple in 1940 London.  ...  {\color{olive} [2]} Robert toughens up on sea duty and in time becomes a petty officer.  {\color{olive} [3]} His hands are badly burned when his ship is sunk, but he stoically rows in the lifeboat for five days without complaint. {\color{olive} [4]} He recuperates in a hospital, tended by {\color{red} \bf Elena}, a beautiful nurse.  {\color{olive} [5]} He is attracted to {\color{rel} \bf \texttt{her}}, but {\color{rel} \bf \texttt{she}} informs him that {\color{rel} \bf \texttt{she}} lost her beloved husband only six months earlier, kisses him, and leaves. \\

\midrule

(4) Q: Who would have been able to win the tournament with one more round? \\    
Paragraph: {\color{olive} [1]} At a jousting tournament in 14th-century Europe, young squires {\color{blue}\bf \textit{William}} Thatcher, Roland, and Wat discover that their master, Sir {\color{red} \bf Ector}, has died. {\color{olive} [2]} If {\color{rel} \bf \texttt{he}} had completed one final pass {\color{rel} \bf \texttt{he}} would have won the tournament. {\color{olive} [3]} Destitute, {\color{blue}\bf \textit{William}} wears {\color{red} \bf Ector}'s armour to impersonate him, winning the tournament and taking the prize.  \\

    \bottomrule
    \end{tabular}
    \end{center}
    \caption{ Examples from QUOREEF~\citep{QUOREF} that were correctly predicted by CorefBERT\BASESIZE, but wrongly predicted by BERT\BASESIZE. {\color{blue}\bf \textit{Answers from BERT\BASESIZE}}, {\color{red} \bf Answers from CorefBERT\BASESIZE}, and {\color{rel} \bf \texttt{Clue}} are colored respectively. }

    \label{tab:case_study}
\end{table}
\section{Case Study on QUOREF}

Table \ref{tab:case_study} shows examples from QUOREF~\citep{QUOREF}. For the first example, it is essential to obtain the fact that the asthmatic boy in question refers to Barry. After that, we should synthesize information from two Mr. Lee's mentions: (1) Mr. Lee trains Barray; (2) Mr. Lee is the uncle of Noreen. Reasoning over the above information,  we could know that Noreen's uncle trains the asthmatic boy. For the second example, it needs to infer that Tippett is a composer from the second sentence for obtaining the final answer from  the first sentence. After training on the mention reference prediction task, CorefBERT has become capable of reasoning over these mentions, summarizing messages from mentions in different positions, and finally figuring out the correct answer.
For the third and fourth examples, it is necessary to know \emph{she} refers to Elena, and \emph{he} refers to Ector by respective coreference resolution. Benefiting from a large amount of distant-supervised coreference resolution training data, CorefBERT successfully finds out the reference relationship and provides accurate answers.

\begin{table}[!t]
\begin{center}

\begin{tabular}{p {0.46\textwidth}}
\toprule
\textbf{Eclipse (Meyer novel)} \\
{\color{olive} [1]} {\color{blue}\bf \textit{Eclipse}} is the third novel in the {\color{red} \bf Twilight Saga} by {\color{red} \bf Stephenie Meyer}. It continues the story of Bella Swan and her vampire love, {\color{blue}\bf \textit{Edward Cullen}}. 
{\color{olive} [2]}  The novel explores Bella's compromise between her love for {\color{blue}\bf \textit{Edward}} and her friendship with shape-shifter {\color{blue}\bf \textit{Jacob Black}},   ... 
{\color{olive} [3]} {\color{blue}\bf \textit{Eclipse}} is preceded by {\color{blue}\bf \textit{New Moon}} and followed by {\color{blue}\bf \textit{Breaking Dawn}}. 
{\color{olive} [4]} The book was released on {\color{red} \bf August 7, 2007}, with an initial print run of one million copies,  and sold more than 150,000 copies in the first 24 hours alone. \\
\midrule
\textbf{Subject}:\,\, {\color{blue}\bf \textit{New Moon}} / {\color{blue}\bf \textit{Breaking Dawn}}   \\
\textbf{Object}: \;  {\color{red} \bf Twilight Saga}  \\
\textbf{Relation}:  {\color{rel} \bf \texttt{Part of the series}}  \\
\midrule
\textbf{Subject}: \,\,{\color{blue}\bf \textit{Edward Cullen}} / {\color{blue}\bf \textit{Jacob Black}}  \\
\textbf{Object}: \;  {\color{red} \bf Stephenie Meyer} \\
\textbf{Relation}:   {\color{rel} \bf \texttt{Creator}} \\
\midrule
\textbf{Subject}:\,\, {\color{blue}\bf \textit{Eclipse}} \\
\textbf{Object}: \; {\color{red} \bf August 7, 2007} \\
\textbf{Relation}: {\color{rel} \bf \texttt{Publication date}} \\
\bottomrule
\end{tabular}
\end{center}
\caption{An example from DocRED~\citep{DocRED}. We show some relational facts detected by CorefBERT\BASESIZE but missed by BERT\BASESIZE.}

\label{tab:docred_case_study}

\end{table}

\section{Case Study on DocRED}
Table \ref{tab:docred_case_study} shows an example from DocRED~\citep{DocRED}.  We show some relational facts detected by CorefBERT\BASESIZE but missed by BERT\BASESIZE. 
For the first relational fact, it is necessary to connect the first and the third sentences through the co-reference of Eclipse for acquiring the fact that New Moon and Breaking Dawn are also the novel in the Twilight Saga. 
For the second and the third relational fact, the referring expressions \emph{it}, \emph{the novel}, and \emph{the book} should be linked to {Eclipse} correctly to increase model's confidence to find out all the characters and the publication date of the novel from the context. CorefBERT considers coreference information of text, which helps to discover relation facts beyond sentence boundary.

\begin{table}[!t]
\begin{center}

\begin{tabular}{p {0.46\textwidth}}
\toprule
\textbf{Claim}: {\color{blue}\bf \textit{Bob Ross}} created {\color{rel} \bf \texttt{ABC}} drama {\color{red} \bf The Joy of Painting}. \\
\midrule
\vspace{-0.7em}
{\color{olive} [1]  \bf  {[Bob Ross]}}
{\color{blue}\bf \textit{Robert Norman Ross}}  was an American painter and television host. \\
\vspace{-0.7em}
{\color{olive} [2]  \bf  [Bob Ross]}
{\color{blue}\bf \textit{He}} was the creator and host of {\color{red} \bf The Joy of Painting}, an instructional television program that aired from 1983 to 1994 on {\color{rel} \bf \texttt{PBS}} in the United States, and also aired in Canada, ... \\
\vspace{-0.7em}
{\color{olive} [3]  \bf  {[Bob Ross]} }
  {\color{red} \bf The Joy of Painting}  is an American half hour instructional television show hosted by painter {\color{blue}\bf \textit{Bob Ross}}  which ran from January 11, 1983, until May 17, 1994. \\
\vspace{-0.7em}
{\color{olive} [4] \bf  [The Joy of Painting]}
 In each episode, {\color{blue}\bf \textit{Ross}}   taught techniques for landscape oil painting, completing a painting in each session. \\
\vspace{-0.7em}
{\color{olive} [5]  \bf  [The Joy of Painting]}
 The program followed the same format as its predecessor, The Magic of Oil Painting , hosted by {\color{blue}\bf \textit{Ross}}'s mentor. \\ 
\midrule
\textbf{Label}: REFUTES \\
\bottomrule
\end{tabular}
\end{center}
\caption{An example from FEVER~\citep{Fever}. Five pieces of evidence from article {\color{olive}  \bf  [Bob Ross]} and  {\color{olive}  \bf  [The Joy of Painting]} are retrieved by the retriever. }
\label{tab:fever_case_study}

\end{table}

\section{Case Study on FEVER}
Table \ref{tab:fever_case_study} shows an example from FEVER~\citep{Fever}. The given claim is fabricated since the drama ``The Joy of Painting" was aired on PBS instead of  ABC. With the CorefBERT encoder, KGAT~\citep{Zhenghao} could propagate and aggregate the entity information from evidence for refuting the wrong claim more accurately. 

\section{Task-Specific Model Details}

All the models are implemented based on Huggingface transformers\footnote{https://github.com/huggingface/transformers}. We train models on downstream tasks with Adam optimizer~\citep{Adam}.

\subsection{Question Answering (QA)}

For QA models, we uses a batch size of $32$ instances with a maximum sequence length of $512$. 

We adopt the official data split for QUOREF~\citep{QUOREF}, where train / development / test set contains 19399 / 2418 / 2537 instances respectively. And we submit our model to the test sever\footnote{https://leaderboard.allenai.org/quoref/submissions/public} for online evaluation.  We conduct a grid search on the learning rate ($lr$) in  $[1\times10^{-5}, 2\times10^{-5}, 3\times10^{-5}]$ and epoch number in $[2, 4, 6]$. The best BERT\BASESIZE configuration on development set used $lr=2\times10^{-5}$, $6$ epochs.  We adopt this configuration for the BERT\LARGESIZE and RoBERTa\LARGESIZE models. 
We regard MRQA~\citep{MRQA} as a testbed to examine whether models can answer questions well across various data distributions. For fair comparison, we keep $lr=3\times10^{-5}$, $2$ epochs for all of the MRQA experiments.

For TASE~\citep{TASE} with CorefRoBERTa encoder, we keep the same configuration\footnote{https://github.com/eladsegal/tag-based-multi-span-extraction} as that of the original paper, which used a batch size of $12$, learning rate of $5\times10^{-6}$, $35$ epochs.

\subsection{Document-level Relation Extraction}

We modify the official code\footnote{https://github.com/thunlp/DocRED} to implement BERT-based models for DocRED~\citep{DocRED}. In our implementation, the representation of a mention, which consists of several words, is the average of  representations of those words. Furthermore, the representation of an entity is defined as the mean of all  mentions referring to it. Finally, two entities' representations are fed to a bi-linear layer to predict relations between them.

We use the official data split for DocRED, where train / development / test set consists of 3053 / 1000 / 1000 documents respectively.  We adopt batch size of $32$ instances with maximum sequence length of $512$ and conduct a grid search on the learning rate in $[2\times10^{-5}, 3\times10^{-5}, 4\times10^{-5}, 5\times10^{-5}]$ and number epochs in $[100, 150, 200]$. We find the  configuration used learning rate of $4\times10^{-5}$, $200$ epochs is best for both the base and the large model. We evaluate models on development set every $5$ epochs and save the checkpoint with the highest F1 score. After that, the test results of the best model are submitted to the evaluation server\footnote{https://competitions.codalab.org/competitions/20717}.

\subsection{Fact Extraction and Verification}

We apply the released code\footnote{https://github.com/thunlp/KernelGAT} of KGAT~\citep{Zhenghao} for evaluating CorefBERT.  We use the official data split for FEVER~\citep{Fever}, where train / development / test set contains 145449 / 19998 / 19998 claims respectively. We adopt a batch size of $32$, maximum length of $512$ tokens and search the learning rate in $[2\times10^{-5},3\times10^{-5},5\times10^{-5}]$. We achieved  the best performance  with learning rate of $5\times10^{-5}$ for the base model and $2\times10^{-5}$ for the large model. All models are trained with a batch size of $32$ instances for $3$ epochs and evaluated on development set every $1000$ steps. After that, we submit test results of our best model to evaluation server\footnote{https://competitions.codalab.org/competitions/18814}. 

\begin{table*}[!t]
\small
\centering
\begin{tabular}{l c c c c c c c c}
\toprule
Model & MNLI & QQP & QNLI & SST-2 & CoLA &  STS-B & MRPC & RTE \\
\midrule
CorefBERT\BASESIZE & $2\times10^{-5}$ &  $4\times10^{-5}$ & $3\times10^{-5}$&  $3\times10^{-5}$  & $5\times10^{-5}$ &  $4\times10^{-5}$&  $5\times10^{-5}$&  $4\times10^{-5}$\\
CorefBERT\LARGESIZE &  $2\times10^{-5}$ & $2\times10^{-5}$ & $2\times10^{-5}$ & $2\times10^{-5}$  &  $3\times10^{-5}$ & $5\times10^{-5}$ & $5\times10^{-5}$ & $3\times10^{-5}$\\
\bottomrule
\end{tabular}
\caption{Learning rate for CorefBERT on GLUE benchmarks.}
\label{gluelr}
\end{table*}

\begin{table*}[!t]
\small
\centering
\begin{tabular}{l r r r r r r r}
\toprule
Model  & Parameters & Layers & Hidden & Embedding & Vocabulary\\
\midrule
CorefBERT\BASESIZE & 110M & 12 & 768& 768&  28,996\\
CorefBERT\LARGESIZE & 340M & 24 & 1,024& 1,024&  28,996\\
CorefRoBERTa\LARGESIZE & 355M & 24 & 1,024& 1,024& 50,265 &\\
\bottomrule
\end{tabular}
\caption{Parameter number and the configuration of CorefBERT.}
\label{number_of_parameters}
\end{table*}

\begin{table*}[!t]
\small
\centering
\begin{tabular}{l c c c c c c c}
\toprule
Model  & QUOREF & MRQA & DocRED & FEVER & GLUE & Coref.\\
\midrule
CorefBERT\BASESIZE & $13.23$  &  $13.15$  & $117.37$ & $18.88$ & $2.95$ & $4.27$\\
CorefBERT\LARGESIZE & $43.40$  & $43.37$  & $180.65$& $54.03$ & $9.22$ & $10.90$\\
\bottomrule
\end{tabular}
\caption{Average inference runtime per example for CorefBERTs on different benchmarks. Inference is done on a RTX 2080ti GPU with a batch of 32 instances and inference time is measured in milliseconds.  The input sequence length is 512 for QUOREF, MRQA, DocRED, FEVER, and 128 for others. Coref.: Coreference resolution. }
\label{average_runtime}
\end{table*}

\subsection{Coreference Resolution}

We use the released code\footnote{https://github.com/vid-koci/bert-commonsense} of WikiCREM~\citep{wikicrem} for fine-tuning BERT-LM~\citep{ASimpleMethodforCommonsenseReasoning} and \CorefBERT on supervised datasets. For a sentence $S$, which possesses a correct candidate $\mathbf{a}$ and an incorrect candidate $\mathbf{b}$, the loss consists of two parts: (1) the negative log-likelihood of the correct candidate; (2) a max-margin between the log-likelihood of the correct candidate and the incorrect candidate:
\begin{equation}
\begin{split}
    \mathcal{L} &= - \log \Pr(\mathbf{a}|S) \\
    & + \alpha\max\left(0, \log \Pr(\mathbf{b}|S) - \log \Pr(\mathbf{a}|S) + \beta\right), \\
\end{split}
\end{equation}
where $\alpha, \beta$ are hyperparameters. We follow the data split and fine-tuning setting of WikiCREM, which adopts a batch size of 64, a  maximum sequence length of $128$ and $10$ epochs training. We search the learning rate $lr \in [3\times10^{-5}, 1\times10^{-5}, 5\times10^{-6}, 3\times10^{-6}]$, hyperparameters $\alpha\in [5,10,20]$, $\beta \in [0.1,0.2,0.4]$. The best performance of models with base size and CorefBERT\LARGESIZE on validation set were achieved with $lr=3\times10^{-5}$, $\alpha=10$, $\beta=0.2$. We keep this configuration for the RoBERTa-based models.

\subsection{Generalized Language Understanding (GLUE)}

We evaluate \CorefBERT on the main GLUE benchmark~\citep{GLUE} used in BERT, including MNLI~\citep{MNLI}, QQP\footnote{https://www.quora.com/q/quoradata/First-Quora-Dataset-Release-Question-Pairs}, QNLI~\citep{rajpurkar2016squad}, SST-2~\citep{SST-2}, CoLA~\citep{CoLA} ,  STS-B~\citep{STS-B}, MRPC~\citep{MRPC} and RTE~\citep{RTE}.

We use a batch size of $32$, maximum sequence length of 128,  fine-tune models for $3$ epochs for all GLUE tasks and select the learning rate of Adam among $[2\times10^{-5}, 3\times10^{-5}, 4\times10^{-5}, 5\times10^{-5}]$ for the best performance on the development set. After that, we submit the result of our best model to the official evaluation server\footnote{https://gluebenchmark.com}. Table~\ref{gluelr} shows the best learning rate  for CorefBERT\BASESIZE and CorefBERT\LARGESIZE.

\subsection{Number of Parameters and Average Runtime}

CorefBERT's architecture is a multi-layer bidirectional Transformer~\citep{Transformer}. Tables~\ref{number_of_parameters} shows the  parameter number of CorefBERTs with different model size. Compared to BERT~\citep{BERT}, \CorefBERT add a few parameters for computing the copy-based objective. Hence, \CorefBERT keeps similar number of parameters as BERT with the same size.

Table~\ref{average_runtime} shows the task-specific average inference runtime per example for CorefBERT. The inferenece is done on a RTX 2080ti GPU with a batch of 32 instances. The inference time includes time on CPU and time on GPU. CorefRoBERTa\LARGESIZE consumes a similar time as CorefBERT\LARGESIZE since they both use a 24-layer Transformer architecture.

\subsection{Resolving the Coreference in the Corpus}
In our preliminary experiment, we resolve the coreference of training corpus via the StanfordNLP tool\footnote{{https://stanfordnlp.github.io/CoreNLP}} and apply our copy-based objective on this training corpus. We find the obtained model performs better than the BERT model without NSP but worse than the current CorefBERT. We think that considering coreference such as pronoun in pre-training could also enhance model's coreferential reasoning ability, while how to deal with the noise from coreference tools remains a problem to be explored.

\end{document}